# Continual Event Recognition for Autonomous Optical Network Monitoring

Khouloud Abdelli
Advanced Photonics Department
Nokia Bell Labs
Stuttgart, Germany
khouloud.abdelli@nokia.com

*Abstract*—**Autonomous optical networks require monitoring systems capable of adapting to evolving disturbances without repeated retraining or service interruption. We propose a continual event recognition framework for state-of-polarization (SOP)-based optical monitoring based on prototype representations, open-set recognition, and few-shot prototype registration. The framework supports recognition of known disturbances, detection and integration of novel events, and investigates whether overlapping disturbances can be inferred through compositional reasoning from existing prototypes. Experimental validation on a 2600-km ROADM-based optical testbed demonstrates 99.82% closed-set recognition accuracy, successful adaptation to novel disturbances without encoder retraining, and reveals that overlapping optical disturbances exhibit varying degrees of latent compositionality. While some overlaps can be recognized in a zero-shot manner through prototype composition, others require explicit few-shot registration. These results establish continual event recognition as a promising paradigm for future autonomous optical network monitoring systems.**

## I. INTRODUCTION

Future optical networks are expected to evolve from passive communication infrastructures into autonomous systems capable of self-monitoring, self-diagnosis, self-optimization, and self-healing with minimal human intervention [1]. Achieving this vision requires monitoring frameworks that continuously adapt to changing operating conditions, emerging failure modes, and previously unseen disturbances without interrupting network operation.

State-of-polarization (SOP) monitoring has recently emerged as a promising sensing modality for optical network monitoring because it enables the detection of mechanical and environmental perturbations using existing coherent transponders without requiring dedicated sensing hardware. Combined with machine learning, SOP measurements have enabled accurate recognition of disturbances such as bending, vibration, impacts, and fiber movements [2-4].

However, most existing approaches operate under a closed-set assumption in which all disturbance classes are known during training and remain fixed after deployment. In practice, optical networks continuously encounter new events arising from environmental changes, infrastructure aging, or previously unobserved failure mechanisms. Conventional classifiers either misclassify such events with high confidence or require costly data collection and complete model retraining whenever new disturbances emerge. This raises a fundamental question:

*How can optical monitoring systems recognize and integrate previously unseen disturbances after deployment without retraining or service interruption?*

A second challenge arises from overlapping disturbances generated by multiple simultaneous perturbations acting on the same fiber segment. Existing monitoring systems typically treat these events as independent predefined classes, requiring explicit training data for every possible combination. Since the number of overlaps grows combinatorially with the number of monitored disturbances, exhaustive data collection rapidly becomes impractical. This motivates a second question:

*Can previously unseen overlapping disturbances be inferred directly from previously learned event representations, or do they require explicit registration? More fundamentally, do overlapping optical disturbances exhibit compositional behavior in the learned representation space?*

These challenges motivate a transition from conventional closed-set classification toward **continual event recognition**, in which monitoring systems continuously recognize known disturbances, detect novel events, and incrementally incorporate new knowledge after deployment.

In this work, we propose a continual event recognition framework for SOP-based optical monitoring based on prototype representations, open-set recognition, and few-shot prototype registration. The proposed framework enables recognition of known disturbances, adaptation to previously unseen events using only a few support samples, and investigates whether novel overlapping disturbances can be recognized through prototype composition or require explicit registration. Experimental validation is performed on a 2600-km ROADM-based optical testbed using SOP measurements collected under four single and four overlapping disturbances generated by robotic actuators. The results reveal that overlapping optical disturbances exhibit varying degrees of latent compositionality: some overlaps support accurate zero-shot recognition through prototype composition, whereas others require explicit few-shot registration. To the best of our knowledge, this represents the first experimental investigation of continual and compositional event recognition for optical network monitoring.

## II. CONTINUAL EVENT RECOGNITION FRAMEWORK

Figure 1 illustrates the proposed continual event recognition framework for autonomous optical network monitoring. Unlike conventional closed-set monitoring systems that assume a fixed set of disturbances throughout deployment, the proposed framework continuously evolves after deployment by recognizing known disturbances, detecting novel events, and incrementally incorporating new knowledge without retraining the underlying encoder.

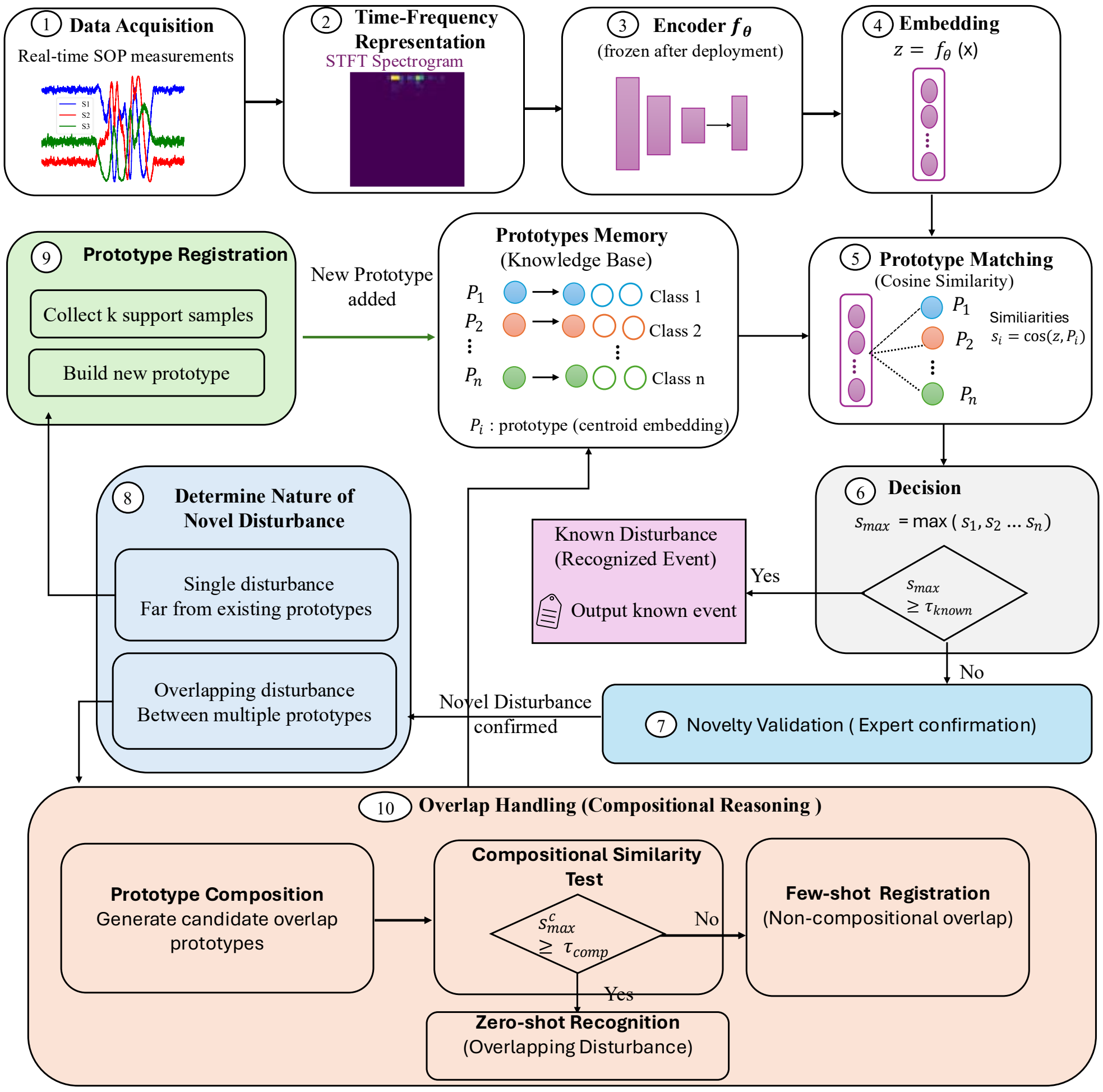

Fig. 1. Proposed prototype-based continual learning framework for SOP monitoring. Known disturbances are recognized through prototype matching, while novel events are incorporated through few-shot prototype registration or overlap composition without retraining the encoder.

The workflow begins with real-time SOP measurements collected by coherent transponders integrated within the optical network. The Stokes parameters are segmented into fixed-length temporal windows and transformed into time-frequency representations using the Short-Time Fourier Transform (STFT). The resulting spectrograms are subsequently projected by an encoder $f_\theta$ into a normalized latent embedding space according to

$$z = f_\theta(x).$$

The encoder is trained only once during the initial deployment stage using the known disturbance classes and remains frozen throughout subsequent operation.

Recognition is performed through prototype matching rather than a conventional softmax classifier. A prototype memory stores representative embeddings corresponding to previously learned disturbances, where each prototype is computed as the centroid of the embeddings belonging to a particular class. For an incoming observation, cosine similarities between the embedding and all stored prototypes are computed and the maximum similarity score

$$s_{\max} = \max_i \cos(z, P_i)$$

is compared against an acceptance threshold $\tau_{\text{known}}$. If $s_{\max} \geq \tau_{\text{known}}$, the event is recognized as a known disturbance and the corresponding label is returned. Otherwise, the observation is treated as an open-set candidate and routed to a novelty validation stage involving expert inspection, maintenance records, operator feedback, or additional measurements. Once the novelty of the disturbance has been confirmed, the framework determines whether the event corresponds to a previously unseen single disturbance or an overlapping disturbance involving multiple known events.

For novel single disturbances, a small support set containing $K$ labeled observations is collected and projected into the latent space using the frozen encoder. A new prototype is constructed by averaging the support embeddings and inserted directly into the prototype memory, enabling immediate recognition of future occurrences without retraining.

For overlapping disturbances, the framework investigates whether the event can be explained through previously acquired knowledge. Candidate overlap prototypes are generated by composing existing disturbance prototypes and compared with the observed embedding through a compositional similarity score $s^c$ . If the maximum compositional similarity satisfies

$$s_{\max} \geq \tau_{\text{comp}},$$

the overlap is recognized in a zero-shot manner through prototype composition. Otherwise, the overlap is treated as a genuinely novel event and incorporated through the same few-shot registration mechanism used for single disturbances.

Consequently, the framework dynamically selects between zero-shot recognition and incremental adaptation depending on the degree of compositionality exhibited by the observed disturbance while keeping the encoder parameters fixed throughout operation.

## III. Experimental Validation

### A. Experimental Setup

Experimental validation is performed using the long-haul SOP monitoring testbed illustrated in Fig. 2(a). The setup consists of a 2600-km loop-back optical link interconnecting multiple fiber spans through reconfigurable optical add-drop multiplexers (ROADMs). Raman-based optical line amplifiers (OLAs) are deployed along the link to compensate for transmission losses and maintain signal integrity over the propagation distance.

State-of-polarization monitoring functionality integrated within the coherent transponder continuously records the three normalized Stokes parameters at millisecond resolution. Mechanical disturbances are generated using Arduino-controlled robotic actuators positioned along the optical link. Four elementary disturbances are considered in this work, namely fiber bending, shaking, small impacts, and vertical displacement. To emulate realistic operating conditions involving multiple simultaneous impairments, additional overlapping disturbances are generated by synchronizing pairs of actuators acting concurrently on the fiber.

To increase dataset diversity and improve robustness to varying operating conditions, a polarization scrambler randomizes the initial SOP state prior to each acquisition. Individual disturbances are generated using a single actuator, whereas overlapping events are produced by simultaneously activating two actuators with temporal offsets varying between 0.5 s and 4 s, thereby generating diverse overlap patterns and interaction dynamics.

Following acquisition, the recorded SOP trajectories are preprocessed to isolate the disturbance interval while removing the polarization transients introduced by the scrambler. Fixed-length windows containing 1000 synchronized samples of the three normalized Stokes parameters are subsequently extracted and transformed into STFT spectrograms for continual event recognition. Representative examples of the acquired SOP trajectories for both single and overlapping disturbances are illustrated in Fig. 2(b).

### B. Continual Event Recognition Protocol

To evaluate the proposed framework under realistic deployment conditions, we design a continual event recognition protocol that emulates the operational lifecycle of an autonomous optical monitoring system after deployment. Unlike conventional closed-set evaluation protocols, the proposed setting assumes that new disturbances and new combinations of disturbances progressively emerge over time and must be incorporated without retraining the underlying encoder. The protocol consists of **three sequential phases**.

#### 1) Phase I: Initial Deployment

The monitoring system is initially trained using only the four elementary disturbances, namely bending (C1), shaking (C2), impacts (C3), and vertical displacement (C4). The encoder and the corresponding class prototypes are learned exclusively from these disturbances and subsequently frozen.

#### 2) Phase II: Novel Single-Event Adaptation

Following deployment, the system is exposed to a previously unseen fan-induced perturbation that mimics a flappy patch-cord event [5] and is intentionally excluded from the initial training stage. The framework first detects the event through open-set recognition and subsequently incorporates it through prototype registration using a small number of support samples while keeping the encoder fixed.

#### 3) Phase III: Continual Adaptation to Overlapping Disturbances

The framework is subsequently exposed to previously unseen overlapping disturbances generated by simultaneous perturbations. Four composite disturbances are considered: $C_5 = C_1 + C_2$, $C_6 = C_1 + C_3$, $C_7 = C_1 + C_4$, and $C_8 = C_2 + C_3$. The overlapping-event experiments investigate three progressively more challenging adaptation scenarios:

1. Can an encoder trained exclusively on single disturbances generalize to overlapping events?

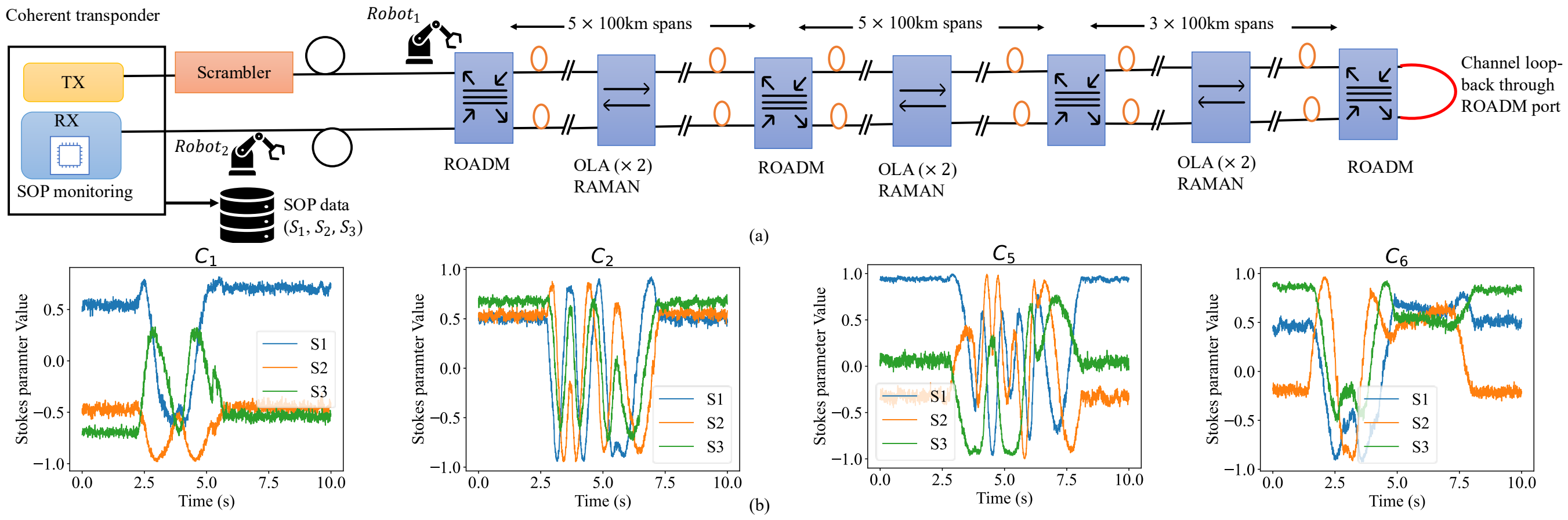

Figure 2. Experimental setup and representative SOP responses used for optical fiber event classification. (a) Long-haul ROADM loop testbed with coherent transceiver, SOP monitoring, and robotic perturbation generation. (b) Example Stokes parameter trajectories corresponding to representative fiber events.

2. Can previously observed overlaps facilitate recognition of unseen overlap combinations?

3. Can unseen overlaps be recognized directly through prototype composition, or do they require few-shot registration?

### C. *Representation Learning Configuration*

The proposed continual event recognition framework relies on a fixed encoder that maps SOP observations into a discriminative latent representation space suitable for prototype-based recognition and continual adaptation. Each event window is converted into a three-channel log-magnitude STFT spectrogram corresponding to the normalized Stokes parameters $S_1$, $S_2$, and $S_3$, resulting in an input tensor of size $3 \times 65 \times 28$ . The spectrogram is processed by a convolutional encoder composed of successive convolutional blocks with batch normalization, GELU activation, channel-attention mechanisms, and pooling operations, followed by a projection head generating an $L_2$ -normalized embedding vector of dimension $d = 160$. The encoder is trained exclusively during the initial deployment stage using the elementary disturbances available at deployment. Representation learning is formulated as a metric-learning problem combining an ArcFace angular-margin loss, a supervised contrastive loss, and a center loss,

$$\mathcal{L} = \lambda_1 \mathcal{L}_{Arc} + \lambda_2 \mathcal{L}_{Con} + \lambda_3 \mathcal{L}_{Cen},$$

which jointly encourage compact intra-class clusters and large inter-class separation within the embedding space.

Training is performed using the training partition of the initial deployment dataset, while hyperparameters and decision thresholds are selected using the validation set. Following convergence, the classification head is discarded and only the encoder is retained during deployment. Importantly, the encoder remains frozen throughout all subsequent experiments, including open-set detection, novel event registration, and overlapping-event adaptation. Consequently, all continual adaptation capabilities arise exclusively from prototype manipulation rather than parameter updates, providing a rigorous evaluation of continual event recognition under realistic deployment conditions. The learned embedding space subsequently serves as the foundation for prototype construction, open-set recognition, incremental registration, and prototype composition throughout the continual learning process.

## IV. Results and Discussion

### A. *Closed-Set Recognition Performance*

Before evaluating continual adaptation capabilities, the quality of the learned latent representation is first assessed under conventional closed-set conditions using the four elementary disturbances available during the initial deployment stage, namely bending (Event 1), shaking (Event 2), small impact (Event 3), and vertical displacement (Event 4).

Recognition is performed through prototype matching in the learned embedding space without retraining or fine-tuning during inference. As shown in Fig. 3(a), the proposed framework achieves an overall classification accuracy of 99.82% and a macro-F1 score of 99.82%, with only a single misclassification observed between Event 2 and Event 4. The confusion matrix therefore demonstrates excellent separability between the investigated SOP disturbances.

To further analyze the learned representation, Fig. 3(b) visualizes the latent embedding space using t-SNE. The four disturbance classes form compact and well-separated clusters, while the corresponding class prototypes are located near the centers of their associated distributions. This observation confirms that the metric-learning objectives successfully minimize intra-class variability while maximizing inter-class separation, producing a structured embedding space suitable for prototype-based recognition, open-set detection, and continual adaptation.

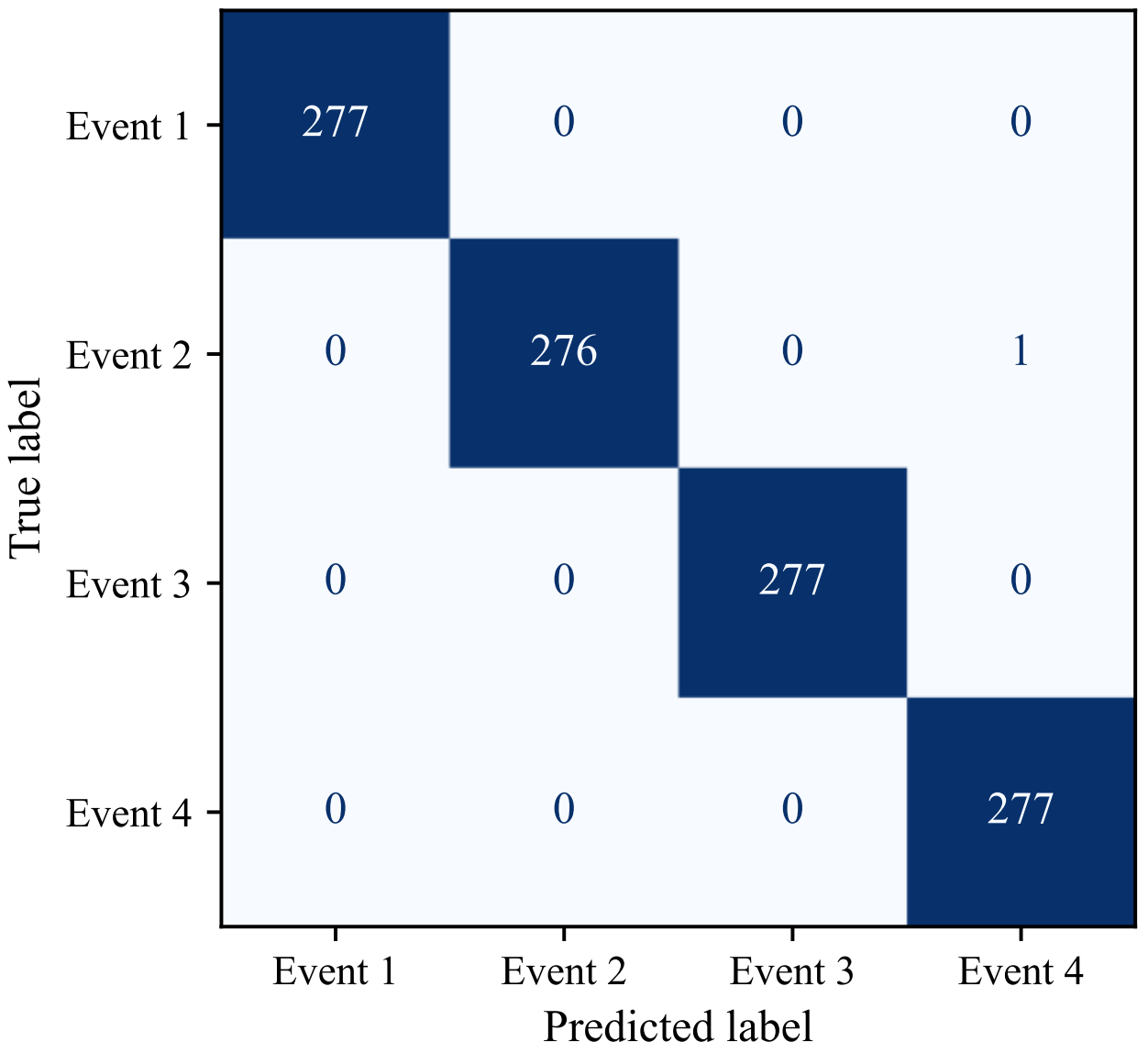

Fig. 3(a). Confusion matrix for closed-set recognition of the four elementary SOP disturbances using prototype-based classification in the learned embedding space.

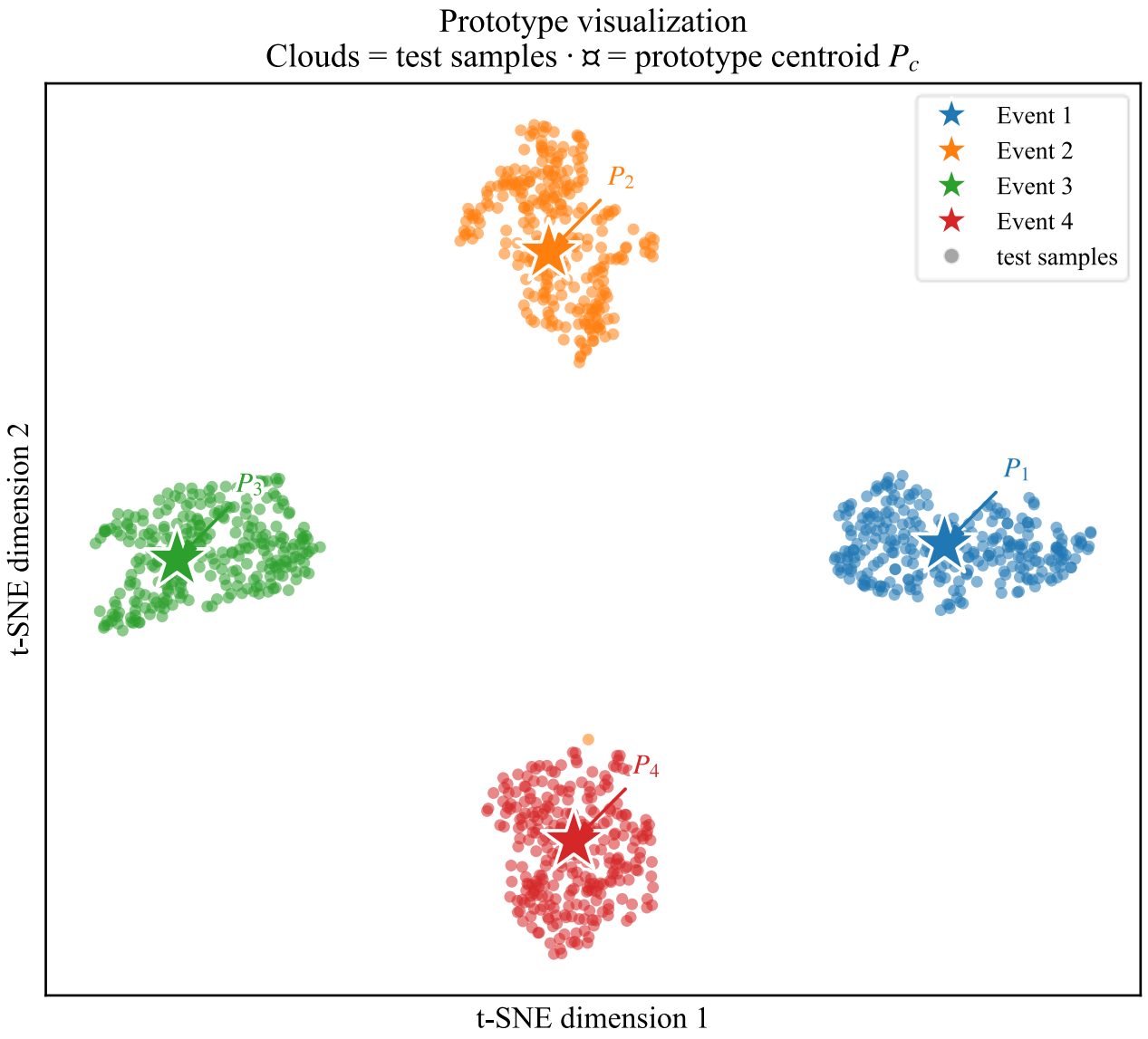

Fig. 3(b). t-SNE visualization of the learned latent representation space. Colored clouds denote test embeddings, while star markers indicate the corresponding class prototypes obtained as centroid embeddings.

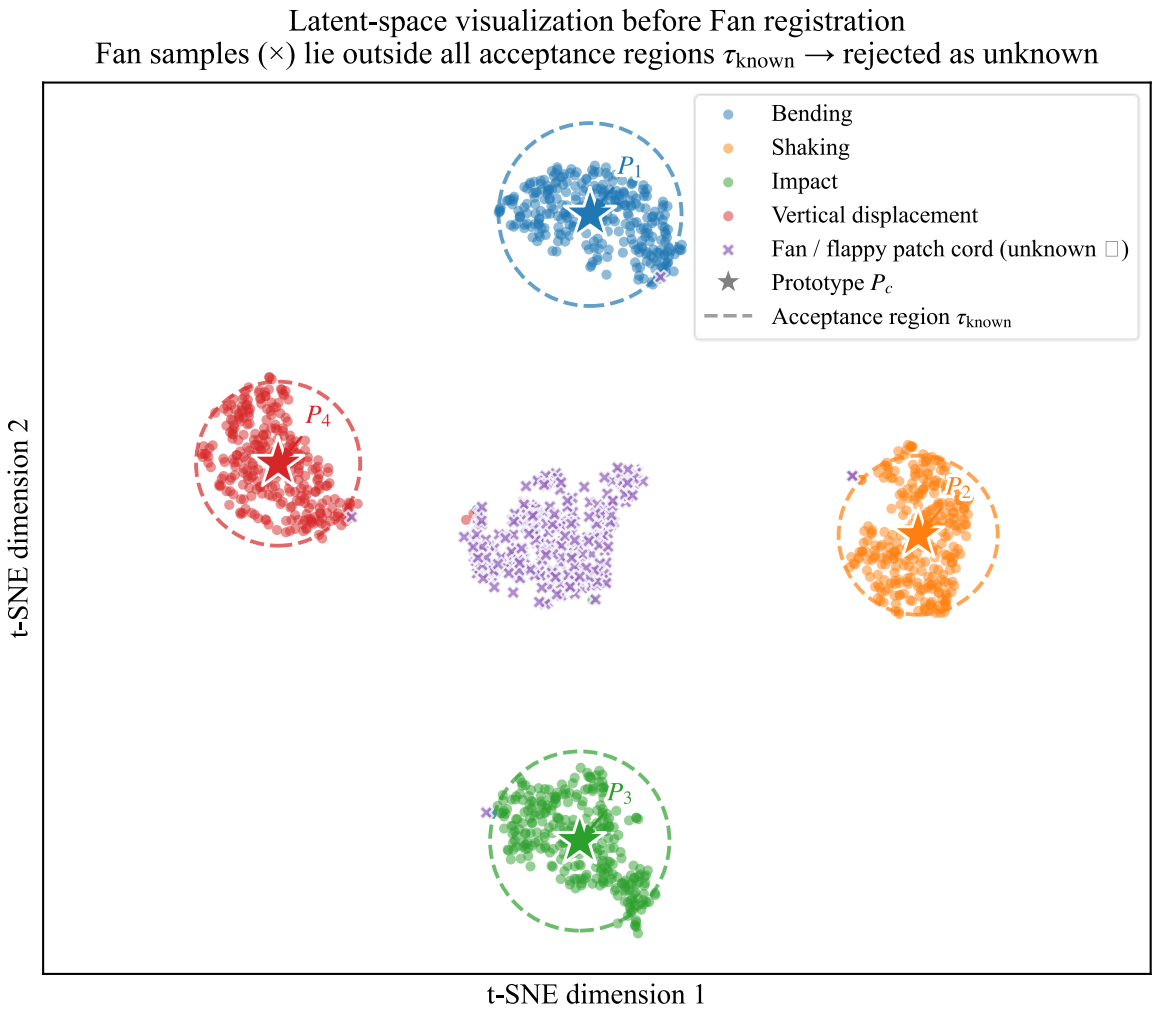

Fig. 4(a). t-SNE visualization of the latent representation before registration of the fan-induced disturbance. The novel event forms a distinct cluster separated from the known disturbances and is therefore rejected through open-set recognition.

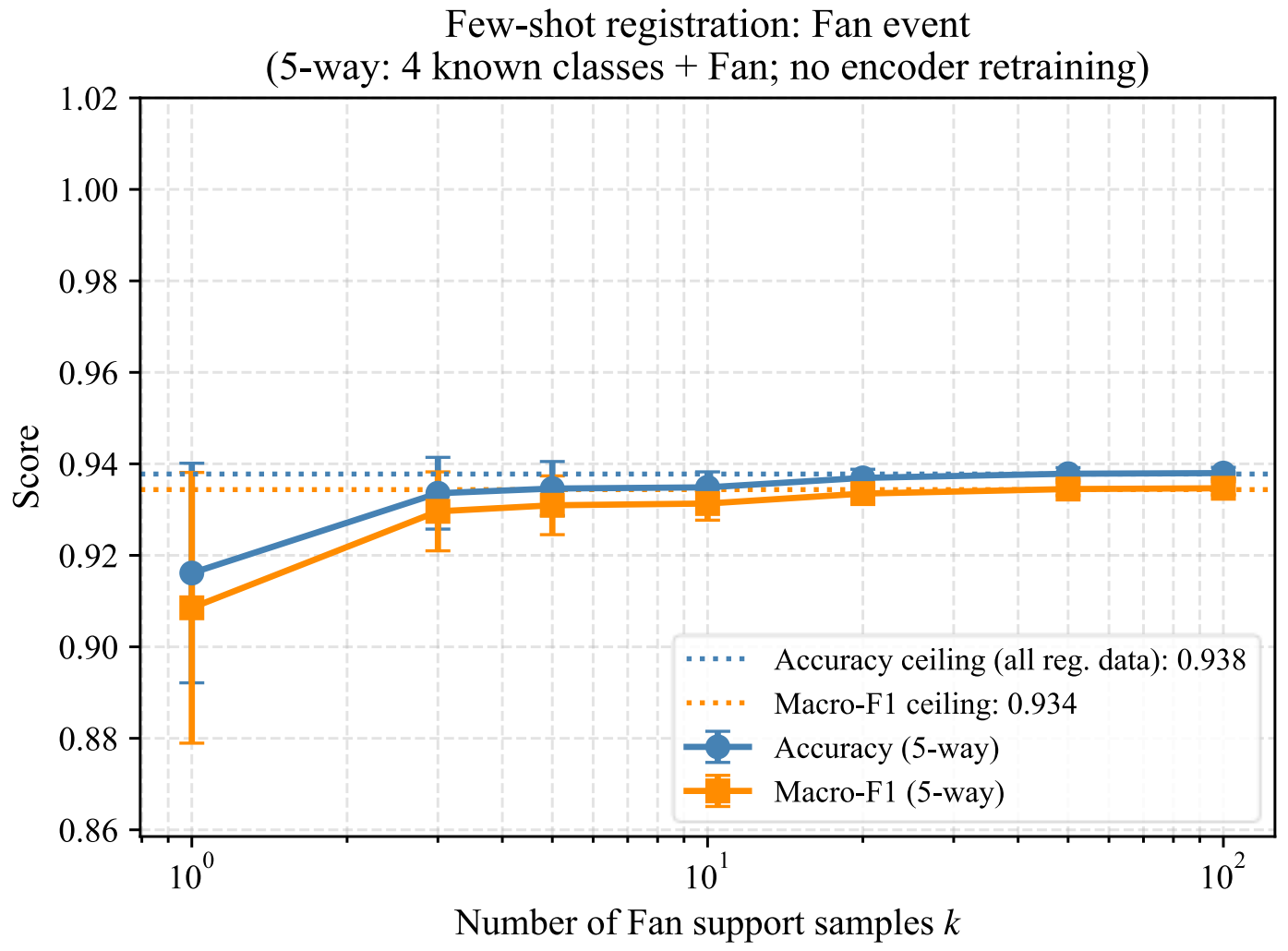

Fig. 4(b). Few-shot registration performance for the fan-induced disturbance as a function of the number of support samples used to construct the new prototype.

## B. *Continual Adaptation to a Novel Single Disturbance*

To evaluate post-deployment adaptation, a previously unseen fan-induced perturbation designed to emulate a flappy patch-cord event is introduced after the initial deployment stage. The fan disturbance is intentionally excluded from training and is acquired using an independent SOP monitoring setup with a different link configuration, thereby introducing both a novel disturbance and a domain shift relative to the original operating conditions.

The objective of this experiment is twofold: (i) determine whether the framework can identify the event as unknown through open-set recognition, and (ii) evaluate whether the disturbance can subsequently be incorporated through prototype registration without retraining the encoder.

Figure 4(a) visualizes the latent representation space before registration. The fan observations form a distinct cluster that is clearly separated from the four known disturbance classes and lies outside their corresponding prototype regions. Consequently, the observations are correctly rejected as unknown rather than being incorrectly assigned to an existing class. Using an acceptance threshold of $\tau_{known} = 0.99$, approximately 99.8% of fan samples are successfully identified as unknown, demonstrating reliable open-set detection under deployment conditions.

Once the disturbance has been confirmed as novel, a new prototype is constructed from a small support set and inserted directly into the prototype memory while keeping the encoder fixed. As shown in Fig. 4(b), the proposed framework achieves approximately 91.6% five-way accuracy (macro-F1 ≈ 90.8%) using only a single support sample. Accuracy rises quickly to about 93.3% with three samples and saturates at approximately 93.8% (macro-F1 ≈ 93.4%) for k ≥ 20, matching the ceiling obtained when all registration data are used. Beyond k = 3, additional samples yield only marginal gains, which indicates that the learned representation already provides a compact embedding distribution for the new disturbance.

-
.

## C. *Continual Adaptation to Overlapping Disturbances*

. While the previous experiment considered adaptation to previously unseen single disturbances, practical optical networks frequently experience multiple simultaneous perturbations acting on the same fiber segment. Such overlapping disturbances constitute a considerably more challenging continual learning scenario since their signatures may differ substantially from those of their constituent events and the number of possible combinations grows combinatorially with the number of monitored disturbances.

### 1) Generalization from single disturbances

Figure 5 illustrates the latent representation obtained when the encoder is trained exclusively on isolated disturbances and evaluated on previously unseen overlaps. The overlap embeddings occupy intermediate regions between multiple prototypes and exhibit substantial overlap with the single-event clusters, indicating that representations learned solely from isolated disturbances do not naturally generalize to composite perturbations. This behavior is further quantified in Fig. 6, where the single-only encoder achieves only 5% zero-shot accuracy and remains below 26% accuracy even after few-shot registration, substantially underperforming the composite-aware representation.

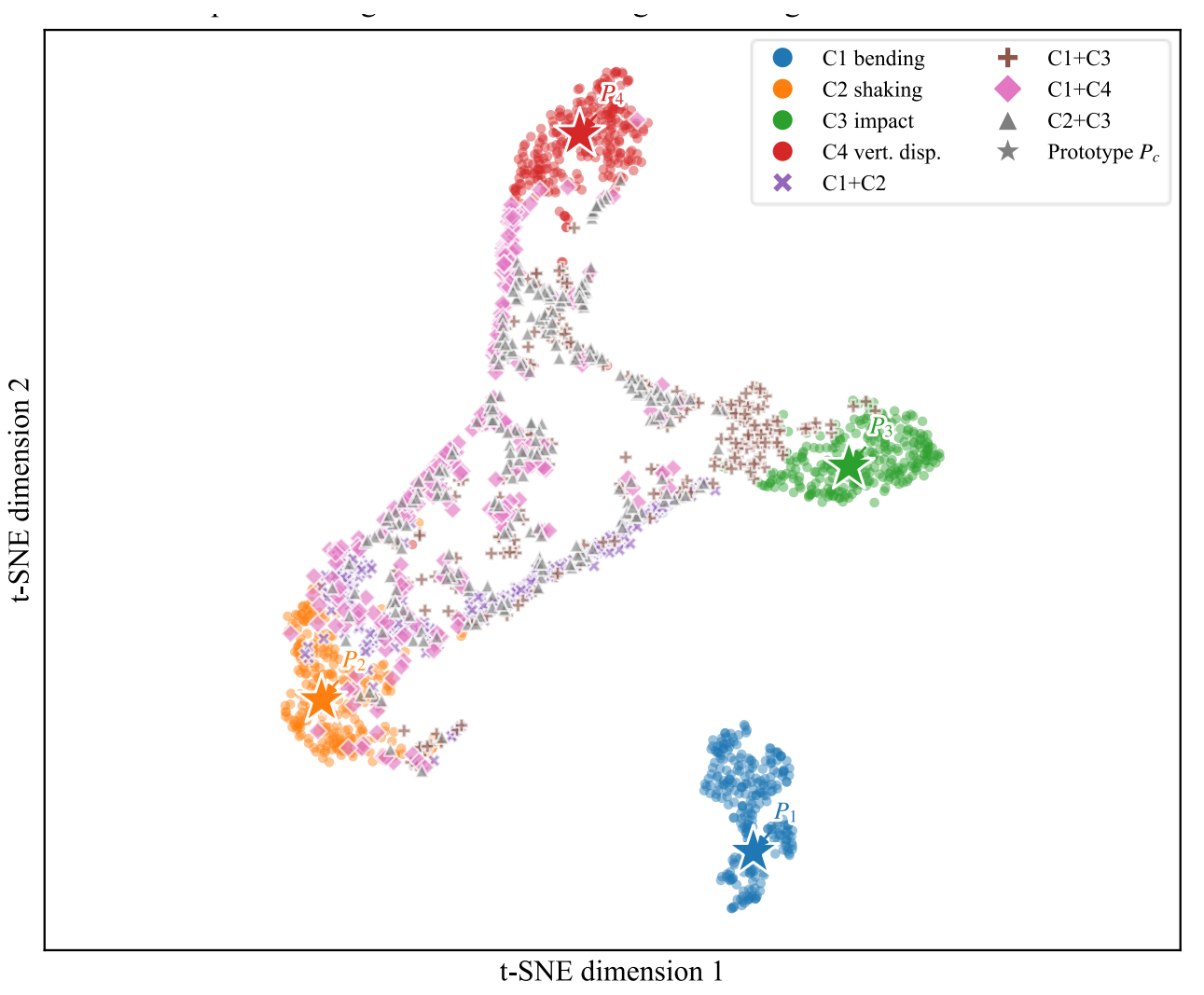

*Fig. 5. t-SNE projection of overlapping disturbances in an embedding space learned from isolated events only.*

### 2) Transfer from previously observed overlaps

A leave-one-overlap-out protocol is subsequently adopted in which the encoder is trained using three overlap classes while one overlap remains unseen during training. As shown in Fig. 6, exposure to a limited number of overlaps dramatically improves generalization to previously unseen combinations, increasing the mean zero-shot accuracy to approximately 71% and exceeding 83% accuracy after observing only a few support samples. The results reveal substantial differences between overlap categories. Certain overlaps, such as C2+C3 and C1+C2, achieve high zero-shot recognition performance despite never being observed during training, whereas others, particularly C1+C4, remain considerably more challenging.

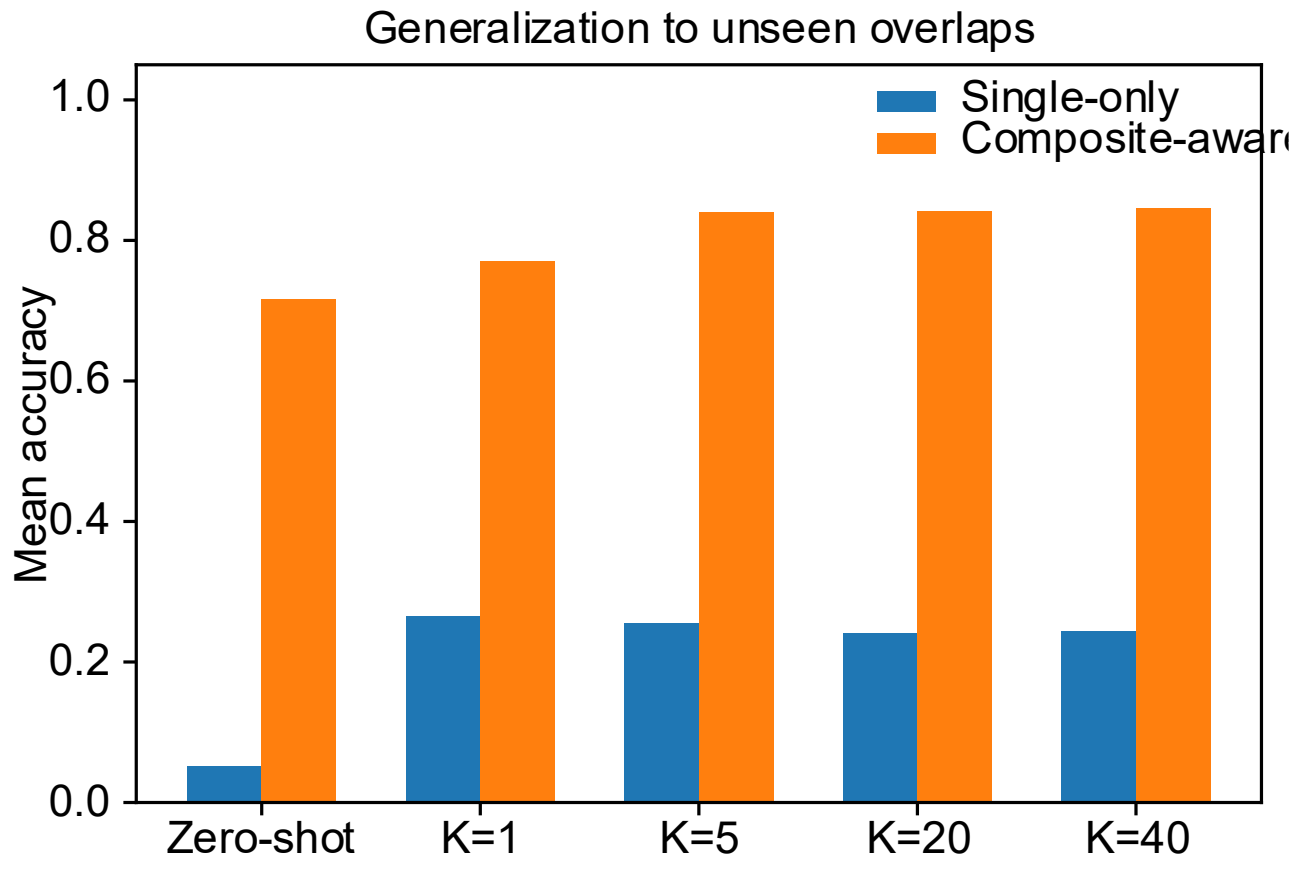

Fig. 6. Generalization to unseen overlaps using single-only and composite-aware representations.

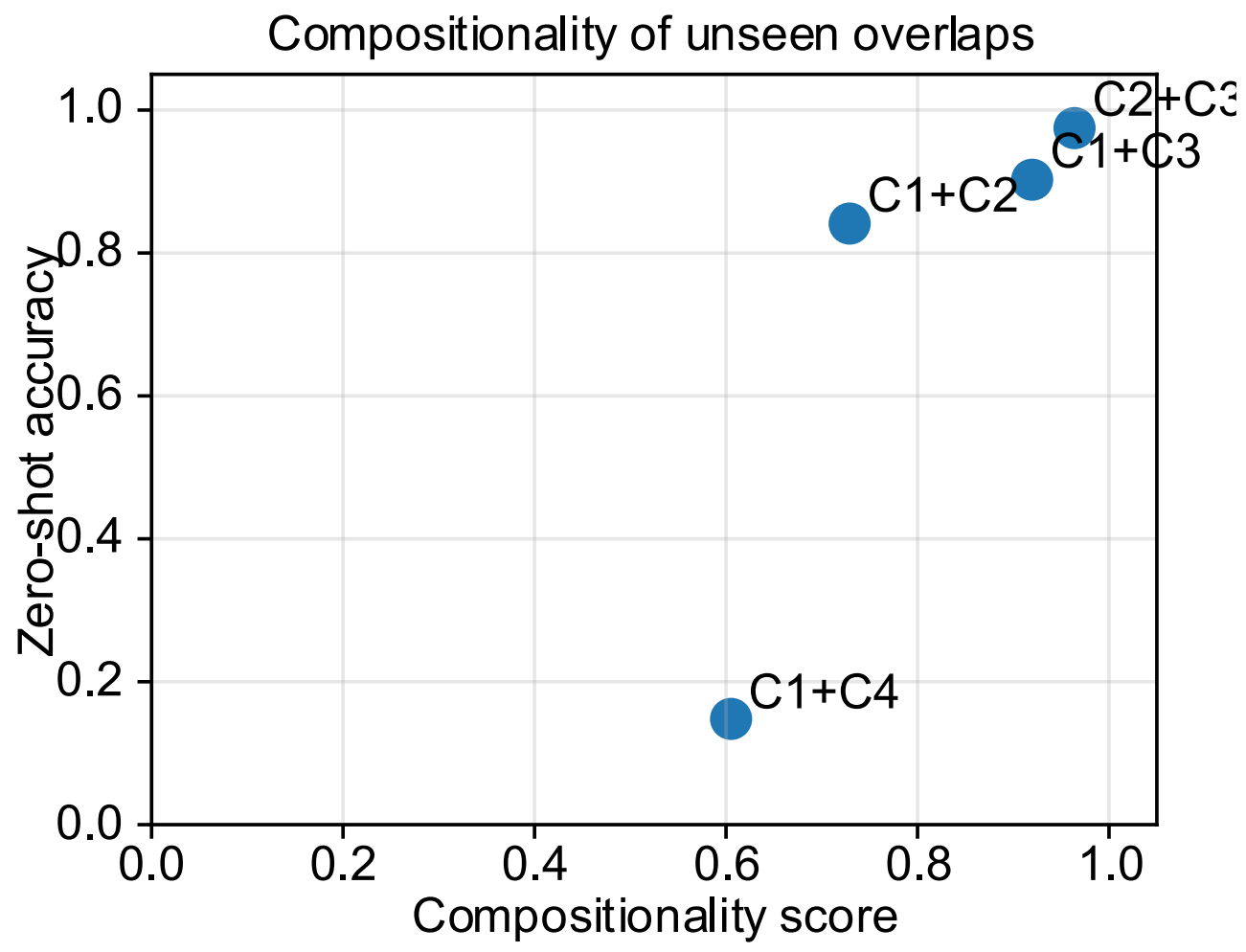

Fig. 7. Compositionality score versus zero-shot recognition accuracy for unseen overlaps.

### 3) Prototype composition and few-shot adaptation

To understand this behavior, the agreement between predicted and observed overlap prototypes is quantified using the compositionality score

$$CS = \cos(P_{ij}^{pred}, P_{ij}^{real}).$$

As shown in Fig. 7, a strong relationship is observed between compositionality score and zero-shot recognition performance. Overlaps exhibiting high compositionality, such as C2+C3 and C1+C2, can be recognized directly from previously acquired knowledge, whereas the weakly compositional overlap C1+C4 exhibits substantially lower zero-shot performance.

For these difficult cases, the framework falls back to few-shot prototype registration. Figure 8 shows that only a small number of support samples are required to approach the fully supervised upper bound. Even for the challenging C1+C4 disturbance, the recognition accuracy increases from approximately 42% with a single support sample to nearly 60% using 40 support samples, while highly compositional overlaps achieve near-ceiling performance with only a few observations. These results demonstrate that previously unseen overlaps may either be recognized directly through prototype composition or incorporated through few-shot registration depending on their degree of compositionality.

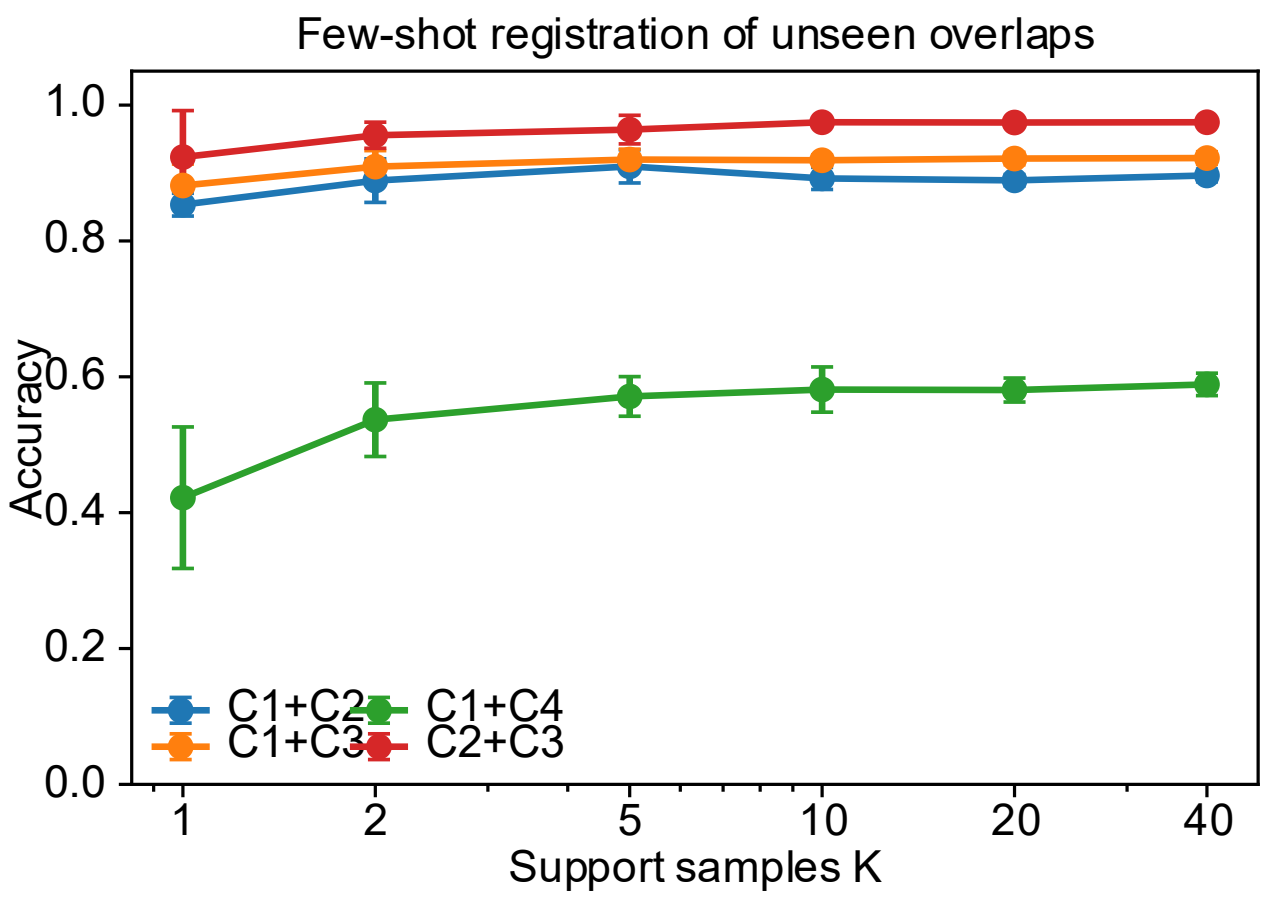

Fig. 8. Few-shot registration performance for previously unseen overlapping disturbances.

## V. CONCLUSIONS

This paper introduced a continual event recognition framework for autonomous optical network monitoring based on prototype representations and open-set adaptation in the latent space of SOP measurements. Experimental validation on a 2600-km long-haul testbed demonstrated 99.82% closed-set recognition accuracy, reliable incorporation of previously unseen disturbances through few-shot prototype registration, and adaptation to overlapping events without encoder retraining. The results further revealed that some overlapping disturbances exhibit latent compositionality and can therefore be recognized directly from previously acquired knowledge, whereas others require explicit registration. These findings establish continual event recognition as a promising paradigm for future self-evolving and self-healing optical networks.

## REFERENCES

[1] K. Abdelli, "Toward Self-Healing Optical Networks: A Vision for AI-Native, Autonomous Resilience," *2025 IPC*, Singapore, 2025, pp. 1-2.

[2] F. Boitier *et al*., “Proactive fiber damage detection in real-time coherent receiver,” in *Proc. ECOC, 2017.*

[3] K. Abdelli *et al*., “Anomaly detection and localization in optical networks using vision transformer and SOP monitoring,” OFC, 2024.

[4] G. Malik *et al*., "Machine Learning for Predictive Multi-Event Detection in Fiber Optic Systems," *ICMLCN*, Barcelona, Spain, 2025.

[5] K. Abdelli, et al., "Breaking boundaries: harnessing unrelated image data for robust risky event classification with scarce state of polarization data," *ECOC, 2023.*